\documentclass{article}
\usepackage{ijcai21}

\usepackage{times}
\usepackage{soul}
\usepackage{url}
\usepackage[hidelinks]{hyperref}
\usepackage[utf8]{inputenc}
\usepackage[small]{caption}
\usepackage{graphicx}
\usepackage{amsmath}
\usepackage{amssymb}
\usepackage{amsthm}
\usepackage{booktabs}
\usepackage{algorithm}
\usepackage{algorithmic}
\usepackage{hyperref}
\usepackage{xspace}

\title{Meta-Reinforcement Learning by Tracking Task Non-stationarity}

\author{
Riccardo Poiani$^1$\footnote{Contact Author}\and
Andrea Tirinzoni$^2$\footnote{Work done while at Politecnico di Milano.}\And
Marcello Restelli$^1$\\
\affiliations
$^1$Politecnico di Milano,\\
$^2$Inria Lille\\
\emails
riccardo.poiani@mail.polimi.it, andrea.tirinzoni@inria.fr, marcello.restelli@polimi.it
}

\newcommand{\algo}{\textsc{TRIO}\xspace}

\newcommand{\wt}[1]{\widetilde{#1}}
\newcommand{\wh}[1]{\widehat{#1}}

\newcommand{\R}{\mathbb{R}}
\newcommand{\E}{\mathbb{E}}

\newcommand{\argmin}{\operatornamewithlimits{argmin}}
\newcommand{\argmax}{\operatornamewithlimits{argmax}}

\renewcommand{\S}{\mathcal{S}} 
\newcommand{\A}{\mathcal{A}} 
\newcommand{\M}{\mathcal{M}} 

\usepackage[colorinlistoftodos, textwidth=30mm]{todonotes}

\begin{document}

\maketitle


\begin{abstract}
Many real-world domains are subject to a structured non-stationarity which affects the agent's goals and the environmental dynamics. Meta-reinforcement learning (RL) has been shown successful for training agents that quickly adapt to related tasks. However, most of the existing meta-RL algorithms for non-stationary domains either make strong assumptions on the task generation process or require sampling from it at training time. In this paper, we propose a novel algorithm (TRIO) that optimizes for the future by explicitly tracking the task evolution through time. At training time, TRIO learns a variational module to quickly identify latent parameters from experience samples. This module is learned jointly with an optimal exploration policy that takes task uncertainty into account. At test time, TRIO tracks the evolution of the latent parameters online, hence reducing the uncertainty over future tasks and obtaining fast adaptation through the meta-learned policy. Unlike most existing methods, TRIO does not assume Markovian task-evolution processes, it does not require information about the non-stationarity at training time, and it captures complex changes undergoing in the environment. We evaluate our algorithm on different simulated problems and show it outperforms competitive baselines.
\end{abstract}


\section{Introduction}

The ability to generalize and quickly adapt to non-stationary environments, where the dynamics and rewards might change through time, is a key component towards building lifelong reinforcement learning (RL) \cite{sutton2018reinforcement} agents. In real domains, the evolution of these environments is often governed by underlying structural and \emph{temporal patterns}. Consider, for instance, a mobile robot navigating an outdoor environment where the terrain conditions are subject to seasonal evolution due to climate change; or where the robot's actuators become less effective over time, e.g., due to the natural degradation of its joints or to the system running out of power; or where the designer changes its desiderata, e.g., how the robot should trade off high-speed movement and safe navigation. The commonality is some unobserved \emph{latent variable} (e.g., the terrain condition, the joints' friction, etc.) that evolves over time with some \emph{unknown} temporal pattern (e.g., a cyclic or smooth change). In this setting, we would expect a good RL agent to (1) quickly adapt to different realizable tasks and (2) to extrapolate and exploit the temporal structure so as to reduce the uncertainty over, and thus further accelerate adaptation to, future tasks.

Meta-RL has proven a powerful methodology for training agents that quickly adapt to related tasks \cite{duan2016rl,wang2016learning,finn2017modelagnostic,hospedales2020meta}. The common assumption is that tasks are i.i.d. from some unknown distribution from which the agent can sample at training time. This assumption is clearly violated in the lifelong/non-stationary setting, where tasks are temporally correlated. Some attempts have been made to extend meta-RL algorithms to deal with temporally-correlated tasks~\cite{al2018continuous,nagabandi2018deep,clavera2018learning,kaushik2020fast,kamienny2020learning,xie2020deep}. However, current methods to tackle this problem have limitations. Some of them \cite{al2018continuous,xie2020deep} model the task evolution as a Markov chain (i.e., the distribution of the next task depends only on the current one). While this allows capturing some cyclic patterns (like seasonal climate change), it is unable to capture more complex behaviors that are frequent in the real world \cite{padakandla2020survey}. Other works \cite{kamienny2020learning} consider history-dependent task-evolution processes but assume the possibility of sampling them during the training. While this assumption seems more reasonable for cyclic processes, where the agent experiences a ``cycle'' infinite many times, it is difficult to imagine that the agent could sample from the same non-stationary process it will face once deployed. Finally, some works \cite{nagabandi2018deep,clavera2018learning,kaushik2020fast} do not explicitly model the task evolution and only meta-learn a policy for fast adaptation to changes. This makes it difficult to handle task-evolution processes other than what they are trained for. These limitations raise our main question: how can we build agents that are able to extrapolate and exploit complex history-dependent task-evolution processes at test time without prior knowledge about them at training time? 

In this paper, we consider the common piecewise stationary setting \cite{ch2020optimizing,xie2020deep}, where the task remains fixed for a certain number of steps after which it may change. Each task is characterized by an unobserved latent vector of parameters which evolves according to an unknown history-dependent stochastic process. We propose a novel algorithm named \algo (TRacking, Inference, and policy Optimization) that is meta-trained on tasks from the given family drawn from different prior distributions, while inferring the ``right'' prior distribution on future tasks by tracking the non-stationarity entirely at test time. More precisely, \algo meta-trains (1) a variational module to quickly infer a distribution over latent parameters from experience samples of tasks in the given family, and (2) a policy that trades off exploration and exploitation given the task uncertainty produced by the inference module. At test time, \algo uses curve fitting to track the evolution of the latent parameters. This allows computing a prior distribution over future tasks, thus improving the inference from the variational module and the fast adaptation of the meta-learned policy. We report experiments on different domains which confirm that \algo successfully adapts to different unknown non-stationarities at test time, achieving better performance than competitive baselines.


\section{Preliminaries}
We model each task as a Markov decision process (MDP) \cite{puterman} $\mathcal{M}_{\omega}=(\S, \A, \mathcal{R}_\omega, \mathcal{P}_\omega, p_0, \gamma)$, where $\S$ is the state space, $\mathcal{A}$ is the action space, $\mathcal{R}_\omega : \S \times \A \times \S \rightarrow \R$ is the reward function, $\mathcal{P}_\omega : \S \times \A \rightarrow \Delta(\S)$ is the state-transition probability function, $p_0$ is the initial state distribution, and $\gamma \in [0,1]$ is the discount factor. We assume each task to be described by a latent vector of parameters $\omega \in \Omega \subset \R^d$ that governs the rewards and the dynamics of the environment, and we denote by $\mathfrak{M} := \{\M_\omega : \omega \in \Omega\}$ the family of MDPs with this parameterization (i.e., the set of possible tasks that the agent can face). We consider episodic interactions with a sequence of MDPs $\M_{\omega_0}, \M_{\omega_1}, \dots$ from the given family $\mathfrak{M}$, which, as in the common piece-wise stationary setting \cite{xie2020deep}, remain fixed within an episode. The evolution of these tasks (equivalently, of their parameters) is governed by a history-dependent stochastic process $\rho$, such that $\omega_t \sim \rho(\omega_0, \dots, \omega_{t-1})$. The agent interacts with each MDP $\M_{\omega_t}$ for one episode, after which the task changes according to $\rho$. At the beginning of the $t$-th episode, an initial state $s_{t,0}$ is drawn from $p_0$; then, the agent chooses an action $a_{t,0}$, it transitions to a new state $s_{t,1} \sim \mathcal{P}_{\omega_t}(s_{t,0},a_{t,0})$, it receives a reward $r_{t,1} = \mathcal{R}_{\omega_t}(s_{t,0},a_{t,0},s_{t,1})$, and the whole process is repeated for $H_t$ steps.\footnote{The length $H_t$ of the $t$-th episode can be a task-dependent random variable (e.g., the time a terminal state is reached).} The agent chooses its actions by means of a possibly history-dependent policy, $a_{t,h} \sim \pi(\tau_{t,h})$, with $\tau_{t,h} := (s_{t,0}, a_{t,0}, s_{t,1}, r_{t,1}, \dots, s_{t,h})$ denoting a $h$-step trajectory, and the goal is to find a policy that maximizes the expected cumulative reward across the sequence of tasks,
\begin{align}
\argmax_{\pi} \E_{\omega_t \sim \rho}\left[\sum_{t=0}^{T-1} \E\left[\sum_{h=0}^{H_t-1} \gamma^h r_{t,h}\ \big| \M_{\omega_t}, \pi\right]\right].
\end{align}
This setting is conceptually similar to hidden-parameter MDPs \cite{velez2013hidden}, which have been used to model non-stationarity \cite{xie2020deep}, with the difference that we allow complex history-dependent task-generation processes instead of i.i.d. or Markov distributions.
\paragraph{Meta-learning setup.} As usual, two phases take place. In the first phase, called \emph{meta-training}, the agent is trained to solve tasks from the given family $\mathfrak{M}$. In the second phase, namely \emph{meta-testing}, the agent is deployed and its performance is evaluated on a sequence of tasks drawn from $\rho$. As in \cite{humplik2019meta,kamienny2020learning}, we assume that the agent has access to the descriptor $\omega$ of the tasks it faces during training, while this information is not available at test time. More precisely, we suppose that the agent can train on any task $\M_\omega$ (for a chosen parameter $\omega$) in the family $\mathfrak{M}$. These assumptions are motivated by the fact that, in practical applications, the task distribution for meta-training is often under the designer's control \cite{humplik2019meta}. Furthermore, unlike existing works, we assume that the agent has \emph{no knowledge} about the sequence of tasks it will face at test time, i.e., about the generation process $\rho$. This introduces the main challenge, and novelty, of this work: how to extrapolate useful information from the family of tasks $\mathfrak{M}$ at training time so as to build agents that successfully adapt to unknown sequences of tasks at test time.

\section{Method}

Imagine to have an oracle that, only at test time, provides the distribution of the parameters of each task \emph{before} actually interacting with it (i.e., that provides $\rho(\omega_1,\dots,\omega_t)$ before episode $t+1$ begins). How could we exploit this knowledge? Clearly, it would be of little use without an agent that knows how the latent parameters affect the underlying environment and/or how to translate this uncertainty into optimal behavior. Furthermore, this oracle works only at test time, so we cannot meta-train an agent with these capabilities using such information. The basic idea behind \algo is that, although we cannot train on the actual non-stationarity $\rho$, it is possible to prepare the agent to face different levels of task uncertainty (namely different \emph{prior} distributions generating $\omega$) by interacting with the given family $\mathfrak{M}$, so as to adapt to the actual process provided by the oracle at test time. More precisely, \algo simulates possible priors from a family of distributions $p_z(\omega) = p(\omega | z)$ parameterized by $z$ and practices on tasks drawn from them. Then, \algo meta learns two components. The first is a module that infers latent variables from observed trajectories, namely that approximates the posterior distribution $p(\omega | \tau, z)$ of the parameters $\omega$ under the prior $p_z$ given a trajectory $\tau$. Second, it meta-learns a policy to perform optimally under tasks with different uncertainty. A particular choice for this policy is a model whose input is augmented with the posterior distribution over parameters computed by the inference module. This resembles a Bayes-optimal policy and allows trading off between gathering information to reduce task uncertainty and exploiting past knowledge to maximize rewards. 

At test time, the two models computed in the training phase can be readily used in combination with $\rho$ (which replaces the simulated priors) to quickly adapt to each task. Obviously, in practice we do not have access to the oracle that we imagined earlier. The second simple intuition behind \algo is that the process $\rho$ can be tracked entirely at test time by resorting to curve fitting. In fact, after completing the $t$-th test episode, the inference model outputs an approximate posterior distribution of the latent parameter $\omega_t$. This, in combination with past predictions, can be used to fit a model that approximates the distribution of the latent variables $\omega_{t+1}$ at the next episode, which in turn can be used as the new prior for the inference model when learning the future task.

Formally, \algo meta-trains two modules represented by deep neural networks: (1) an inference model $q_\phi(\tau,z)$, parameterized by $\phi$, that approximates the posterior distribution $p(\omega | \tau, z)$, and (2) a policy $\pi_\theta(s, q_\phi)$, parameterized by $\theta$, that chooses actions given states and distributions over latent parameters. At test time, \algo learns a model $f(t)$ that approximates $\rho(\omega_0,\dots,\omega_{t-1})$, namely the distribution over the $t$-th latent parameter given the previous ones. We now describe each of these components in detail, while the pseudo-code of \algo can be found in Algorithm~\ref{alg:train} and~\ref{alg:test}.




\begin{algorithm}[t]
\caption{\algo (meta-training)} \label{alg:train}
\begin{algorithmic}[1]
\REQUIRE Task family $\mathfrak{M}$, hyperprior $p(z)$, batch size $n$
\STATE{Randomly initialize $\theta$ and $\phi$}
\WHILE{not done} 
	\STATE{Sample prior parameters $\{z_i\}_{i=1}^n$ from $p(z)$}
	\STATE{Sample task parameters $\{\omega_i\}_{i=1}^n$ from $\{p_{z_i}(\omega)\}_{i=1}^n$}
	\STATE{Collect $\{\tau_i\}_{i=1}^n$ using policy $\pi_{\theta}$ in MDPs $\{\M_{\omega_i}\}_{i=1}^n$}
	\STATE{Update $\theta$ by optimizing \eqref{eq:policy-obj} using $\{\tau_i\}_{i=1}^n$}
	\STATE{Update $\phi$ by optimizing \eqref{eq:elbo} using $\{z_i,\omega_i,\tau_i\}_{i=1}^n$}
\ENDWHILE
\ENSURE Meta-policy $\pi_\theta$ and inference network $q_\phi$
\vspace{0.1cm}
\end{algorithmic}
\end{algorithm}


\begin{algorithm}[t]
\caption{\algo (meta-testing)} \label{alg:test}
\begin{algorithmic}[1]
\REQUIRE Meta-policy $\pi_{\theta}$, inference network $q_{\phi}$, stream of tasks $\omega_{t} \sim \rho$,
initial prior parameters $\wh{z}_0$
\STATE{Initialize $D_{\omega} = \emptyset$}
\FOR{$t=0,1,\dots$}
	\STATE{Interact with $\M_{\omega_t}$ using $\pi_{\theta}$, $q_{\phi}$, $\wh{z}_t$ and collect $\tau_t$}
	\STATE{Predict $\wh{\omega}_t$ using $q_\phi(\tau_t,\wh{z}_t)$ and set $D_{\omega} = D_{\omega} \cup \{\wh{\omega}_{t}\}$}
	\STATE{Fit Gaussian processes using $D_{\omega}$ and predict $\wh{z}_{t+1}$}
\ENDFOR
\vspace{0.1cm}
\end{algorithmic}
\end{algorithm} 

\subsection{Task Inference}

As mentioned before, the inference module aims at approximating the posterior distribution $p(\omega | \tau, z)$ of the latent variable $\omega$ given a trajectory $\tau$ and the prior's parameter $z$.
Clearly, computing the exact posterior distribution $p(\omega | \tau, z) \propto p(\tau | \omega)p_z(\omega)$ is not possible since the likelihood $p(\tau | \omega)$ depends on the true models of the environment $\mathcal{P}_\omega$ and $\mathcal{R}_\omega$, which are unknown. Even if these models were known, computing $p(\omega|\tau,z)$ requires marginalizing over the latent space, which would be intractable in most cases of practical interest. 
A common principled solution is variational inference \cite{Blei_2017}, which approximates $p(\omega|\tau,z)$ with a family of tractable distributions. A convenient choice is the family of multivariate Gaussian distributions over the latent space $\mathbb{R}^d$ with independent components (i.e., with diagonal covariance matrix). Suppose that, at training time, we consider priors $p_z(\omega)$ in this family, i.e., we consider $p_z(\omega) = \mathcal{N}(\mu,\Sigma)$ with parameters $z=(\mu,\sigma)$ given by the mean $\mu\in\R^d$ and variance $\sigma\in\R^d$ vectors, which yield covariance $\Sigma = \mathrm{diag}(\sigma)$. Then, we approximate the posterior as $q_\phi(\tau,z) = \mathcal{N}(\mu_\phi(\tau,z), \Sigma_\phi(\tau,z))$, where $\mu_\phi(\tau,z)\in\mathbb{R}^d$ and $\Sigma_\phi(\tau,z) = \mathrm{diag}(\sigma_\phi(\tau,z))$ are the outputs of a recurrent neural network with parameters $\phi$.

To train the inference network $q_\phi$, we consider a \emph{hyperprior} $p(z)$ over the prior's parameters $z$ and directly minimize the expected Kullback-Leibler (KL) divergence between $q_\phi(\tau,z)$ and the true posterior $p(\omega|\tau,z)$. Using standard tools from variational inference, this can be shown equivalent to minimizing the \emph{evidence lower bound} (ELBO) \cite{Blei_2017},
\begin{align}
\argmin_\phi \E\Big[\E_{\wh{\omega}\sim q_\phi}[\log p(\tau|\wh{\omega},z)] + \mathrm{KL}\big(q_\phi(\tau,z) \big\| p_z\big)\Big],
\end{align}
where the outer expectation is under the joint process $p(\tau,\omega,z)$. In practice, this objective can be approximated by Monte Carlo sampling. More precisely, \algo samples the prior's parameters $z$ from $p(z)$, the latent variable $\omega$ from $p_z(\omega)$, and a trajectory $\tau$ by interacting with $\M_\omega$ under the current policy. Under a suitable likelihood model, this yields the following objective (full derivation in Appendix~\ref{app:elbo}):
\begin{align}
\notag\argmin_{\phi} \sum_{i=1}^n \Big(\|\mu_\phi(\tau_i,z_i) &- \omega_i\|^2 + \mathrm{Tr}(\Sigma_\phi(\tau_i,z_i))\\ &+ \frac{\lambda}{H_i} \mathrm{KL}(q_\phi(\tau_i,z_i) \| p_{z_i}) \Big).\label{eq:elbo}
\end{align}
Here we recognize the contribution of three terms; (1) the first one is the standard mean-square error and requires the mean-function $\mu_\phi(\tau,z)$ to predict well the observed tasks (whose parameter is known at training time); (2) the second term encodes the intuition that this prediction should be the least uncertain possible (i.e., that the variances of each component should be small); (3) the last term forces the approximate posterior to stay close to the prior $p_z(\omega)$, where the closeness is controlled as usual by a tunable parameter $\lambda \geq 0$ and by the length $H_i$ of the $i$-th trajectory. 

\subsection{Policy Optimization}

The agent's policy aims at properly trading off exploration and exploitation under uncertainty on the task's latent parameters $\omega$. In principle, any technique that leverages a given distribution over the latent variables can be used for this purpose. Here we describe two convenient choices.

\paragraph{Bayes-optimal policy.} Similarly to \cite{zintgraf2019varibad}, we model the policy as a deep neural network $\pi_\theta(s, z)$, parametrized by $\theta$, which, given an environment state $s$ and a Gaussian distribution over the task's parameters, produces a distributions over actions. The former distribution is encoded by the vector $z=(\mu,\sigma)$ which is obtained from the prior and refined by the inference network as the data is collected. This policy is meta-trained to directly maximize rewards on the observed trajectories by proximal policy optimization (PPO) \cite{schulman2017proximal},
\begin{align}\label{eq:policy-obj}
\argmax_{\theta} \sum_{i=1}^n \sum_{h=0}^{H_i-1} \gamma^h r_{h,i},
\end{align}
where the sum is over samples obtained through the same process as for the inference module. Similarly to the inference network, the policy is meta-tested without further adaption. Intuitively, being provided with the belief about the task under consideration, this ``Bayes-optimal'' policy automatically trades off between taking actions that allow it to quickly infer the latent parameters (i.e., those that are favorable to the inference network) and taking actions that yield high rewards.

\paragraph{Thompson sampling.} 
Instead of using an uncertainty-aware model, we simply optimize a task-conditioned policy $\pi_\theta(s,\omega)$ to maximize rewards (recall that we have access to $\omega$ at training time). That is, we seek a \emph{multi-task} policy, perhaps one of the most common models adopted in the related literature \cite{lan2019meta,humplik2019meta}. Then, at test time, we can use this policy in combination with Thompson sampling \cite{thompson1933likelihood} (a.k.a. posterior sampling) to trade-off exploration and exploitation in a principled way. That is, we sample some parameter $\omega \sim q_\phi(\tau,z)$ from the posterior computed by the inference network, choose an action according to $\pi_\theta(s,\omega)$, feed the outcome back into the inference network to refine its prediction and repeat this process for the whole episode. As we shall see in our experiments, although simpler than training the Bayes-optimal policy, this approach provides competitive performance in practice.

\subsection{Tracking the Latent Variables}

As we discussed in the previous sections, before interacting with a given task, both the inference network $q_\phi(\tau,z)$ and the policy $\pi_\theta(s,z)$ (assuming that we use the Bayes-optimal model) require as input the parameter $z$ of the prior under which the task's latent variables are generated. While at meta-training we explicitly generate these parameters from the hyperprior $p(z)$, at meta-testing we do not have access to this information. A simple workaround would be to use non-informative priors (e.g., a zero-mean Gaussian with large variance). Unfortunately, this would completely ignore that, at test-time, tasks are sequentially correlated through the unknown process $\rho$. Therefore, we decide to \emph{track} this process online, so that at each episode $t$ we can predict the distribution of the next task in terms of its parameter $\wh{z}_{t+1}$. While many techniques (e.g., for time-series analysis) could be adopted to this purpose, we decide to model $\rho$ as a Gaussian process (GP) \cite{williams2006gaussian} due to its flexibility and ability to compute prediction uncertainty. Formally, at the end of the $t$-th episode, we have access to estimates $\wh{\omega}_0,\dots,\wh{\omega}_{t}$ of the past latent parameters obtained through the inference network \emph{after} facing the corresponding tasks. We use these data to fit a separate GP for each of the $d$ dimensions of the latent variables, while using its prediction one-step ahead $\wh{z}_{t+1} = (\wh{\mu}_{t+1},\wh{\sigma}_{t+1})$ as the prior for the next episode. Intuitively, when $\rho$ is properly tracked, this reduces the uncertainty over future tasks, hence improving both inference and exploration in future episodes.



\section{Related Works}
\textbf{Meta-reinforcement learning}. The earliest approaches to meta-RL make use of \emph{recurrent networks} \cite{hochreiter1997long} to aggregate past experience so as to build an internal representation that helps the agent adapt to multiple tasks \cite{hochreiter2001learning,wang2016learning,duan2016rl}. \emph{Gradient-based} methods, on the other hand, learn a model initialization that can be adapted to new tasks with only a few gradient steps at test time \cite{finn2017modelagnostic,rothfuss2018promp,stadie2018some,liu2019taming}. 
Some approaches of this kind have been used to tackle dynamic scenarios \cite{nagabandi2018deep,clavera2018learning,kaushik2020fast}. \cite{al2018continuous} use few-shot gradient-based methods to adapt to sequences of tasks. Unlike our work, they handle only Markovian task evolution processes and use the knowledge of non-stationarity at training time. Another line of work, which has recently gained considerable attention, considers \emph{context-based} methods that directly take the task uncertainty into account by building and inferring latent representations of the environment. 
\cite{rakelly2019efficient} propose an off-policy algorithm that meta-trains two modules: a variational autoencoder that builds a latent representation of the task the agent is facing, and a task-conditioned optimal policy that, in combination with posterior sampling, enables structured exploration of new tasks. \cite{zintgraf2019varibad} design a similar model, with the main difference that the policy is conditioned on the entire posterior distribution over tasks, thus approximating a Bayes-optimal policy. 
All of these methods are mainly designed for stationary multi-task settings, while our focus is on non-stationary environments. 
For the latter setup, \cite{kamienny2020learning} meta-learn a reward-driven representation of the latent space that is used to condition an optimal policy. Compared to our work, they deal with continuously-changing environments and assume the possibility of ``simulating'' this non-stationarity at training time, an assumption that might be violated in many real settings.

\textbf{Non-stationary reinforcement learning}. Since most real-world applications involve environments that change over time, non-stationary reinforcement learning is constantly gaining attention in the literature (see \cite{padakandla2020survey} for a detailed survey). \cite{xie2020deep} aim at learning dynamics associated with the latent task parameters and perform online inference of these factors. However, their model is limited by the assumption of Markovian inter-task dynamics. Similar ideas can be found in \cite{ch2020optimizing}, where the authors perform curve fitting to predict the agent's return on future tasks so as to prepare their policy for changes in the environment. Here, instead, we use curve fitting to track the evolution of the latent task parameters and we learn a policy conditioned on them. 


\section{Experiments}
Our experiments aim at addressing the following questions:
\begin{itemize}
	\item Does \algo successfully track and anticipate changes in the latent variables governing the problem? How does it perform under different non-stationarities?
	\item  What is the advantage w.r.t. methods that neglect the non-stationarity? How better an oracle that knows the task evolution process can be?
\end{itemize}
To this end, we evaluate the performances of \algo in comparison with the following baselines:
\begin{itemize}
	\item \textbf{Oracles}. At the beginning of each episode, they have access to the correct prior from which the current task is sampled. They represent the best that the proposed method can achieve. 
	\item \textbf{VariBAD} \cite{zintgraf2019varibad} and \textbf{RL2} \cite{wang2016learning}, which allow us to evaluate the gain of tracking 
	non-stationary evolutions w.r.t. inferring the current task from scratch at the beginning of each episode.
	\item \textbf{MAML (Oracle)} \cite{finn2017modelagnostic}. To evaluate gradient-based methods for non-stationary settings, we report the ``oracle'' performance of MAML \emph{post-adaptation} (i.e., after observing multiple rollouts from the current task).
\end{itemize}
Furthermore, we test two versions of our approach: Bayes-\algo, where the algorithm uses the Bayes-optimal policy model, and TS-\algo, where we use a multi-task policy in combination with Thompson sampling. Additional details and further results, can be found in Appendix~\ref{app:experiments}.

\begin{figure*}[t]
\centering
\includegraphics[height=8.8cm]{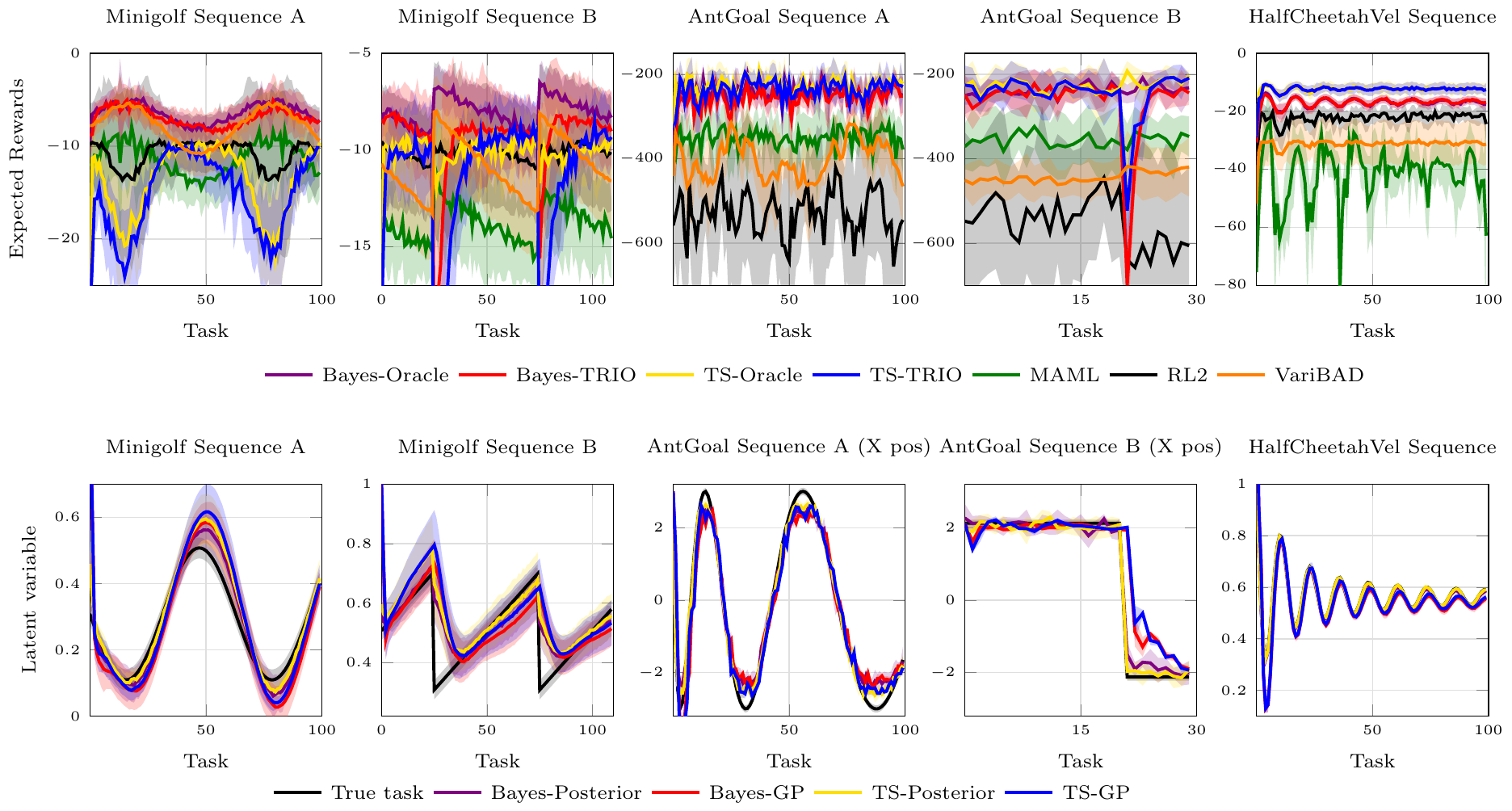} 
\caption{Meta-test performance on different sequences of the selected domains. All plots concerning the Minigolf (MuJoCo) domain are 
averages and standard deviations of 20 (5) policies, each of which is tested 50 times on the same episode; each task is composed of 4 (1) episodes.
\textit{(top)} Expected rewards per task. 
\textit{(bottom)} Latent-variable tracking per task. The figures report the true latent variable of each task (True task), the posterior mean of \algo at the end of each task (Bayes-Posterior and TS-Posterior), and the GP prediction of \algo for the next task (Bayes-GP and TS-GP). For the first task, Bayes-GP and TS-GP are replaced by the initial prior given to the algorithm. For the AntGoal sequences, we report only the x-coordinate of the goal position. Plots for the y-coordinate are very similar and can be found in Appendix~\ref{app:experiments}.}
\label{fig:all}
\end{figure*}


\subsection{Minigolf}

In our first experimental domain, we consider an agent who is playing a minigolf game day after day. In the minigolf domain \cite{tirinzoni2019transfer}, the agent has to shoot a ball, which moves along a level surface, inside a hole with the minimum number of strokes. In particular, given the position of the ball on the track, the agent directly controls the angular velocity of the putter. The reward is $0$ when the ball enters the hole, $-100$ if it goes beyond the hole, and $-1$ otherwise. The problem is non-stationary due to different weather conditions affecting the dynamic friction coefficient of the ground. This, in turn, greatly affects the optimal behavior. Information on how this coefficient changes are unknown a-priori, thus they cannot be used at training time. However, the temporal structure of these changes make them suitable for tracking online.
At test time, we consider two realistic models of the ground's friction non-stationarity: A) a sinusoidal function, which models the possible seasonal behavior due to changing weather conditions; and B) a sawtooth-shaped function, which models a golf course whose conditions deteriorate over time and that is periodically restored by human operators when the quality level drops below a certain threshold. 

\paragraph{Results.} Let us first analyze the tracking of the latent variables in Figure~\ref{fig:all} (\textit{bottom}). As we can see, the proposed algorithms are able to successfully track the a-priori unknown evolution of the friction coefficient in both sequences. 


As shown in Figure~\ref{fig:all} \textit{(top)}, Bayes-\algo achieves the best results in this domain. It is worth noting that its performance overlaps with the one of its oracle variant for the whole sinusoidal task sequence, while in the sawtooth ones performance drops occur only when the golf course gets repaired (i.e., when its friction changes abruptly). Indeed, before these abrupt changes occur, the agent believes that the next task would
have a higher friction and, thus, it strongly shoots the ball towards the goal. However, as the friction abruptly drops, the agent overshoots the hole, thus incurring a highly negative reward. This behavior is avoided in few episodes, when the agent recovers the right evolution of the friction coefficient and restores high performance. A similar reasoning applies to TS-\algo, which, however, obtains much lower performance, especially in sequence A. 
The main cause of this problem is its na\"ive exploration policy and the way TS handles task uncertainty. In fact, since its policy is trained conditioned on the true task, the agent successfully learns how to deal with correct friction values; however, even when negligible errors in the inference procedure are present, the agent incurs catastrophic behaviors when dealing with small friction values and it overshoots the ball beyond the hole. When the friction is greater, as close to the peaks of sequence B, the errors in the inference network have less impact on the resulting optimal policies, and TS-\algo achieves the same performance as Bayes-\algo. \newline
VariBAD achieves high performance in situations of low abrasion, but its expected reward decreases as friction increases. This is due to the fact that, at the beginning of each episode, the agent swings the putter softly seeking information about the current abrasion level. While this turns out to be optimal for small frictions, as soon as the abrasion increases, these initial shots become useless: if the agent knew that a high friction is present, it could shoot stronger from the beginning without risking to overshoot the ball. A similar behavior is observed for RL2. Finally, MAML (Oracle) suffers from worse local maxima than context-based approaches and performs significantly worse.

\subsection{MuJoCo}

We show that \algo successfully scales to more complex problems by evaluating its performance on two MuJoCo benchmarks typically adopted in the meta-RL literature. We consider two different environments: 1) HalfCheetahVel, in which the agent has to run at a certain target velocity; and 2) AntGoal, where the agent needs to reach a certain goal in the 2D space. We modify the usual HalfCheetahVel reward function to make the problem more challenging as follows: together with a control cost, the agent gets as reward the difference between its velocity and the target velocity; however, when this difference is greater than $0.5$, an additional $-10$ is added to model the danger of high errors in the target velocity and to incentivize the agent to reach an acceptable speed in the smallest amount of time. For AntGoal, we consider the typical reward function composed of a control cost, a contact cost, and the distance to the goal position.
At test time, the non-stationarity affects the target speed in HalfCheetahVel and the goal position in AntGoal. We consider different non-linear sequences to show that \algo can track complex temporal patterns. 

\paragraph{Results.} Figure~\ref{fig:all} reports two sequences for AntGoal and one for HalfCheetahVel. As we can see, in all cases, TS-\algo and Bayes-\algo successfully track the changes occurring in the latent space. In HalfCheetahVel, our algorithms outperform state-of-the-art baselines. In this scenario, TS-\algo achieves the best results. We conjecture that this happens due to its simpler task-conditioned policy model, which potentially leads to an easier training process that ends up in a slightly better solution. Interestingly, differently to what reported in \cite{zintgraf2019varibad}, we also found RL2 to perform better than VariBAD. This might be due to the fact that we changed the reward signal, introducing stronger non-linearities. Indeed, VariBAD, which uses a reward decoder to train its inference network, might have problems in reconstructing this new function, leading to a marginally worse solution. Finally, MAML (Oracle) suffers from the same limitation as in the Minigolf domain.

In AntGoal, both our algorithms reach the highest performance. It is worth noting that, in line with the Minigolf domain, when the non-stationarity presents high discontinuities (as in sequence B), \algo suffers a perfomance drop which is resolved in only a handful of episodes. MAML (Oracle) achieves competitive performance with VariBAD; however, we recall that MAML's results are shown post-adaptation, meaning that it has already explored the current task multiple times. Finally, being the problem more complex, RL2, compared to VariBAD, faces more troubles in the optimization procedure, obtaining a worse behavior. 

Finally, it has to be highlighted that, in both problems, our algorithms are able to exploit the temporal patterns present in the non-stationarity affecting the latent variables. \emph{Anticipating} the task evolution before it occurs leads to faster adaptation and higher performance.


\section{Conclusions}

We presented \algo, a novel meta-learning framework to solve non-stationary RL problems using a combination of multi-task learning, curve fitting, and variational inference. Our experimental results show that \algo outperforms state-of-the-art baselines in terms of achieved rewards during sequences of tasks faced at meta-test time, despite having no information on these sequences at training time. Tracking the temporal patterns that govern the evolution of the latent variables makes \algo able to optimize for future tasks and leads to highly-competitive results, thus establishing a strong meta-learning baseline for non-stationary settings. 

Our work opens up interesting directions for future work. For example, we could try to remove the need of task descriptors at training time, e.g., by building and tracking a reward-driven latent structure 
\cite{kamienny2020learning} or a representation to reconstruct future rewards \cite{zintgraf2019varibad}. The main challenge would be to build priors over this learned latent space to be used for training the inference module. 

\bibliographystyle{named}
\bibliography{biblio_short}

\newpage
\onecolumn

\appendix

\section{Derivations}\label{app:elbo}

We provide the detailed steps to derive the objective function optimized by \algo to train the inference network. Recall that the goal is to find the distribution $q_\phi(\tau,z)$ that minimizes the expected KL divergence with respect to the true posterior $p(\omega|\tau,z)$,
\begin{align*}
\argmin_\phi \E\Big[\mathrm{KL}\big(q_\phi(\tau,z) \big\| p(\omega|\tau,z)\big)\Big],
\end{align*}
where the expectation is under the joint process $p(\tau,\omega,z)$. Fix a trajectory $\tau$ and a prior's parameter $z$. Then, by the standard decomposition used in deriving the variational lower bound \cite{Blei_2017},
\begin{align*}
\mathrm{KL}\big(q_\phi(\tau,z) \big\| p(\omega|\tau,z)\big) &= \int_{\Omega} q_\phi(\wh{\omega} | \tau,z) \log \frac{q_\phi(\wh{\omega} | \tau,z)}{p(\wh{\omega}|\tau,z)}\mathrm{d}\wh{\omega}\\ &= \int_{\Omega} q_\phi(\wh{\omega} | \tau,z) \log \frac{q_\phi(\wh{\omega} | \tau,z)p(\tau | z)}{p(\wh{\omega},\tau|z)}\mathrm{d}\wh{\omega}\\ &= \int_{\Omega} q_\phi(\wh{\omega} | \tau,z) \log \frac{q_\phi(\wh{\omega} | \tau,z)}{p(\wh{\omega},\tau|z)}\mathrm{d}\wh{\omega} + \log p(\tau | z)\\ &= \int_{\Omega} q_\phi(\wh{\omega} | \tau,z) \log \frac{q_\phi(\wh{\omega} | \tau,z)}{p(\tau|\wh{\omega},z)p(\wh{\omega}|z)}\mathrm{d}\wh{\omega} + \log p(\tau | z)\\ &= -\int_{\Omega} q_\phi(\wh{\omega} | \tau,z) \log p(\tau|\wh{\omega})\mathrm{d}\wh{\omega} + \int_{\Omega} q_\phi(\wh{\omega} | \tau,z) \log \frac{q_\phi(\wh{\omega} | \tau,z)}{p(\wh{\omega}|z)}\mathrm{d}\wh{\omega} + \log p(\tau | z)\\ &= - \E_{\wh{\omega} \sim q_{\phi}(\tau,z)}\big[\log p(\tau|\wh{\omega})\big] + \mathrm{KL}\big(q_\phi(\tau,z) \big\| p_z(\omega)\big) + \log p(\tau | z).
\end{align*}
Since the log-evidence $\log p(\tau | z)$ does not depend on $\phi$, we have that 
\begin{align*}
\argmin_\phi \E\Big[\mathrm{KL}\big(q_\phi(\tau,z) \big\| p(\omega|\tau,z)\big)\Big] = \argmin_\phi \E\Big[- \E_{\wh{\omega} \sim q_{\phi}(\tau,z)}\big[\log p(\tau|\wh{\omega})\big] + \mathrm{KL}\big(q_\phi(\tau,z) \big\| p_z(\omega)\big)\Big].
\end{align*}
It only remains to specify a suitable likelihood model $p(\tau|\wh{\omega})$. Since the actual likelihood would depend on the true task models $\mathcal{P}_{\wh{\omega}}$ and $\mathcal{R}_{\wh{\omega}}$, which are both unknown, we can use the following common workaround. Since at training time we explicitly sample from the joint process $p(\tau,\omega,z)$, so that the true task parameter $\omega$ that generated trajectory $\tau$ is known, we set
\begin{align*}
p(\tau|\wh{\omega}) = e^{- \Lambda\| \wh{\omega} - \omega \|^2},
\end{align*}
where $\Lambda > 0$ is some parameter. Intuitively, we say that a trajectory $\tau$ is more likely if samples $\wh{\omega}$ from the inference network are close to the ground truth $\omega$. This is also motivated by PAC-Bayesian theory\footnote{See, e.g., ``PAC-Bayesian supervised classification: the thermodynamics of statistical learning'' [Catoni, 2007].}, where the likelihood model is typically inversely proportional to the expected loss of the associated supervised learning algorithm (in our case, the inference network). Moreover, from PAC-Bayesian theory we also known that $\Lambda$ should increase with the amount of data available for fitting the model, which in our case is the length $H$ of trajectory $\tau$. Therefore, we set $\Lambda := \frac{H}{\lambda}$, for a different parameter $\lambda > 0$. By plugging this likelihood model into the derivation above and rescaling by $\Lambda$, we obtain
\begin{align*}
\argmin_\phi \E\Big[\mathrm{KL}\big(q_\phi(\tau,z) \big\| p(\omega|\tau,z)\big)\Big] &= \argmin_\phi \E\Big[\E_{\wh{\omega} \sim q_{\phi}(\tau,z)}\big[\| \wh{\omega} - \omega \|^2\big] + \frac{\lambda}{H}\mathrm{KL}\big(q_\phi(\tau,z) \big\| p_z(\omega)\big)\Big]\\ &= \argmin_\phi \E\Big[\| \mu_\phi(\tau,z) - \omega \|^2 + \mathrm{Tr}(\Sigma_\phi(\tau,z)) + \frac{\lambda}{H}\mathrm{KL}\big(q_\phi(\tau,z) \big\| p_z(\omega)\big)\Big],
\end{align*}
where the last step follows by noting that $\wh{\omega} - \omega$ has distribution $\mathcal{N}(\mu_\phi(\tau,z) - \omega, \Sigma_\phi(\tau,z))$ and using the closed-form of the squared $\ell_2$-norm of a multivariate Gaussian random variable. The objective stated in the paper is obtained by approximating the outer expectation with samples $(\tau_i,\omega_i,z_i)$ from the joint process $p(\tau,\omega,z) = p(\tau|\omega)p_z(\omega)p(z)$.


\section{Additional Details on \algo}\label{app:algorithm}

\subsection{Off-prior Inference Training}


\begin{algorithm}[t]
\caption{\algo (meta-training with off-prior data)} \label{alg:train2}
\begin{algorithmic}[1]
\REQUIRE Task family $\mathfrak{M}$, hyperprior $p(z)$, batch size $n$
\STATE{Randomly initialize $\theta$ and $\phi$}
\WHILE{not done} 
	\STATE{Sample prior parameters $\{z_i\}_{i=1}^n$ and $\{\wt{z}_i\}_{i=1}^n$ from $p(z)$}
	\STATE{Sample task parameters $\{\omega_i\}_{i=1}^n$ from $\{p_{z_i}(\omega)\}_{i=1}^n$}
	\STATE{Collect trajectories $\{\tau_i\}_{i=1}^n$ in MDPs $\{\M_{\omega_i}\}_{i=1}^n$ using policy $\pi_{\theta}$ and the true prior $p_{z_i}$}
	\STATE{Collect trajectories $\{\wt{\tau}_i\}_{i=1}^n$ in MDPs $\{\M_{\omega_i}\}_{i=1}^n$ using policy $\pi_{\theta}$ and the wrong prior $p_{\wt{z}_i}$}
	\STATE{Update $\theta$ by optimizing \eqref{eq:policy-obj} using $\{\tau_i\}_{i=1}^n$ (only on-prior data)}
	\STATE{Update $\phi$ by optimizing \eqref{eq:elbo} using $\{z_i,\omega_i,\tau_i\}_{i=1}^n \cup \{\wt{z}_i,\omega_i,\wt{\tau}_i\}_{i=1}^n$ (both on-prior and off-prior data)}
\ENDWHILE
\ENSURE Meta-policy $\pi_\theta$ and inference network $q_\phi$
\vspace{0.1cm}
\end{algorithmic}
\end{algorithm}

Empirically, we have found better performances if we adopt a combination of on-prior and off-prior data for meta-training the
inference network $q_{\phi}$. Indeed, if we adopted only on-prior data as described in Algorithm \ref{alg:train} (that is, by using the exact prior parameters $z$ sampled from the hyperprior $p(z)$), the inference network would be trained on trajectories collected by a policy $\pi_{\theta}$ whose state is augmented with a posterior distribution computed from the \emph{true} prior. However, at test time, the prior itself is not available and must be estimated. When such estimates are poor (e.g., at the beginning of a meta-test sequence, when GPs use few data points, or when abrupt changes occurs in the latent variables), these errors might affect the inference network, which was trained only on correct priors and thus could produce a wrong posterior. In turn, the wrong posterior is fed into the policy, which might take poor actions. While the predictions could become correct as more data is fed into the inference network (i.e., when the contribution of the prior vanishes), this behavior could still severely affect the agent's behavior in early steps of the episode, and thus its final performance.

For this reason, we combine on-prior data with off-prior data as shown in Algorithm \ref{alg:train2}. The idea is to make the inference network more robust to misspecified priors by training it also with trajectories that are generated by an agent with wrong prior
knowledge. More precisely, we sample ``wrong'' prior parameters $\wt{z}$ alongside the ``true'' ones $z$ from the hyperprior $p(z)$. Then, we sample the task's latent variable $\omega$ from the correct prior $p_z$ and collect (1) a trajectory $\tau$ from $\M_\omega$ using the agent's policy $\pi_\theta$ in combination with the inference network $q_\phi$ that uses the \emph{true} prior parameters $z$, and (2) a trajectory $\wt{\tau}$ using the same process with the \emph{wrong} prior parameters $\wt{z}$. Finally, we optimize the inference network $q_\phi$ using both on-prior and off-prior data (i.e., $\tau$ and $\wt{\tau}$), while we use only on-prior data to optimize the policy $\pi_\theta$.

\subsection{Gaussian Processes for Tracking Latent Variables}

In all our experiments, we use a combination of a squared exponential kernel, a white-noise kernel, and a linear kernel. Formally,
\begin{align}
\mathcal{K}(x_{i}, x_{j}) = c \exp\left(-\frac{d(x_{i}, x_{j})^{2}}{2l^{2}}\right) + w(x_{i}, x_{j}) + \sigma^{2}_{0} + x_{i}x_{j},
\end{align}
where $w(x_{i}, x_{j}) = W$ if $x_{i} = x_{j}$ else $0$. Note that $c$, $W$, $l$ and $\sigma^{2}_{0}$ are the hyper-parameters of the kernel. We fix $W$ to 0.01, while we let the GP automatically tune the other hyper-parameters online by maximum likelihood on the data points observed at test time. We chose this kernel model \emph{before} actually defining the experimental domains in such a way that allows us to capture many trends of practical interest. Notably, it turned out to be robust and effective in all the sequences we meta-tested, despite it was not tuned for those sequences. Indeed, as the experiments show, it is able to capture complex behaviors and adapt its hyper-parameters in a way that allows tracking the evolution of the latent space. Notably, despite its smooth nature, this kernel leads to good performance even on test sequences with abrupt changes. It is worth noting that, for such sequences, our approach can easily be integrated with techniques related to monitoring and control to improve performance even further. For instance, we may combine curve fitting with change-point detection methods to reset the samples used for fitting GPs and recover the same performance we have at the beginning of a meta-test sequence. We consider this an important consequence of the simplicity and generality of the approach.

Finally, we note that GPs are sufficiently flexible to allow the introduction of additional prior knowledge. For instance, if one has knowledge about the family of sequences that are faced at test time (e.g., as in \cite{kamienny2020learning,al2018continuous}), it is possible to directly tune the kernel function and its hyper-parameters to obtain better performace.

\subsection{Relaxing \algo's Assumptions}

\paragraph{Non-stationary setting.} 

We designed our algorithm for episodic interactions with a sequence of MDPs that, as in common piece-wise stationary settings, do not change within the episode. We note that this assumption could be easily relaxed, as \algo can deal with latent variables evolving within the episode with only minor tweaks. This can be achieved by resetting the prior before the episode itself ends. For instance, we can define a fixed episode length $H$ and, every $H$ steps, obtain a new prior from the GPs. Intuitively, this works well as far as the underlying task does not change much in $H$ steps.\footnote{In practice, if the task evolves significantly in short periods of time (e.g., with continuous non-smooth or abrupt changes), there might be no learning algorithm able to track the changes and perform well.} In case predicting the new priors becomes computationally prohibitive as the number of data points used to fit the GPs rapidly grows, we can use a sliding-window approach that selects only the most recent $w$ samples.



\paragraph{Availability of task identifier and hyperprior at training time.}

Since \algo aims at training models that generalize to unknown and off-distribution task sequences, the possibility to train on chosen tasks from the given family while knowing the corresponding latent parameters is a quite important requirement. It is intuitively justified by the fact that, in practice, the designer often controls the training tasks. While this is common in simulated settings \cite{humplik2019meta}, it has been done in real environments as well \cite{clavera2018learning}. In order to relax this assumption, we could take inspiration from recent works build latent representations of the environment dynamics and rewards \cite{rakelly2019efficient,zintgraf2019varibad,xie2020deep}. For instance, if we only had the possibility to interact with tasks sampled from an unknown and uncontrollable distribution, or if we only had access to interaction data with multiple unknown tasks, we could first fit an auto-encoder to embed the environment models into a low-dimensional latent space. Then, we could take the decoder as our environment simulator and apply \algo's meta-training procedure in its original form. Of course, as for all meta-learning algorithms, the resulting performance would highly depend on the training distribution and its difference with respect to the task generation process encountered at test time. It is possible that further adaptation at test time can alleviate this issue. While building these kind of task embeddings is a simple approach, we do not exclude that better solutions exist whose study is an exciting direction for future work.

\section{Additional Details on the Experiments}\label{app:experiments}

\subsection{Domains}

\subsubsection{Minigolf}

We briefly recap how the Minigolf domain works, while referring the reader to \cite{tirinzoni2019transfer} for the detailed description. The goal of the agent is to shoot a ball of a given radius $r$ inside a hole of diameter $D$ in the minimum
number of shots. We consider a one-dimensional scenario in which the agent always hits the ball towards the hole, which will move with constant deceleration $d = \frac{5}{7}g\omega$, where $\omega$ is the dynamic friction coefficient of the ground and $g$ is the gravitational acceleration. 
At each step, the agent selects action $a$ within the range $[1e^{-5}, 10]$ that determines the angular speed $s$ of the putter according to the formula $s = al(1+\epsilon)$, where $l$ is the lenght of the putter and $\epsilon \sim \mathcal{N}(0, 0.3)$. 
The angular velocity of the putter will determine the initial velocity $v_{0} = sl$ of the ball. For each distance $x_{0}$ from the current position to the hole, the agent successfully completes the task, obtaining reward $0$, if $v_{0}$ ranges from $v_{min} = \sqrt{2dx_{0}}$ to 
$v_{max} = \sqrt{(2D-r)^{2} \frac{g}{2r} + v^{2}_{min}}$. In the case in which $v_{0}$ exceeds $v_{max}$ the ball will overcome the hole and the episode ends with a reward of $-100$. On the other hand, if $v_{0} < v_{min}$, the episode goes on, and the agent can try to shoot again from the position $x = x_{0} - \frac{v_{0}^{2}}{2d}$. At the beginning of a trial, the agent starts from a random position $x_{0}$ between $2000$cm and $0$cm far from the hole.

At meta-training time, we generate tasks with friction in the range $[0.01, 2]$. Prior means are sampled uniformly from the range $[-1, 1]$, while variances are sampled uniformly from the range $[0.01, 0.2]$. Once a task has been sampled, it gets rescaled in the task range $[0.01, 2]$.

The meta-test sequences presented in Figure~\ref{fig:all} are defined as:
$\omega_{A}(t) = -0.199 \sin(0.1t) + 0.30845$ and $\omega_{B}(t) = 0.5075 + 0.398 (\frac{t}{50} - \left \lfloor{0.5 + \frac{t}{50}}\right \rfloor )$. At meta-test time, we sample tasks from a normal distribution whose mean is set to $\omega(t)$ rescaled in $[-1, 1]$, and whose variance is $0.001$. In both sequences, we use a wrong initial prior for our algorithms by setting the friction mean to $1$ and the standard deviation to $0.2$ (note that mean has to be rescaled in $[-1, 1]$).

\subsubsection{HalfCheetahVel}

In this domain, the only latent variable is the target velocity. During training, the possible target velocities belong to the interval $[0, 1.5]$. Our algorithms sample uniformly prior means from the range $[-1, 1]$, while variances are sampled uniformly from the range $[0.01, 0.3]$. Once tasks are sampled, they get rescaled in the interval $[0, 1.5]$.

The meta-test sequence for HalfCheetahVel presented in Figure~\ref{fig:all} is defined as 
$v(t)=\frac{3}{16} (-\tanh({\frac{t+5}{16}}) + \sin({\frac{t+5}{2}}) \frac{16}{t+5}) + 0.75$. As in Minigolf, we sample tasks from gaussian distribution with mean a rescaled version of $v(t)$ and variance $0.00001$.
As initial prior, we feed our algorithm with wrong prior knowledge: $\mathcal{N}(1.5, 0.01)$ (note that mean has to be rescaled in $[-1, 1]$).

\subsubsection{AntGoal}

\begin{figure*}[t]
\centering
\includegraphics[height=6.0cm]{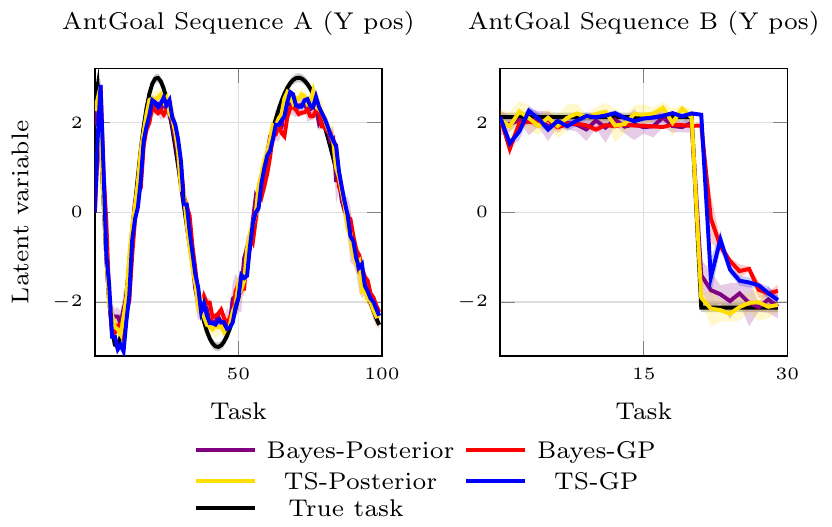}
\caption{Tracking of the Y position in AntGoal sequence A and B.}
\label{fig:anty}
\end{figure*}

\begin{figure*}[t]
\centering
\includegraphics[height=6.0cm]{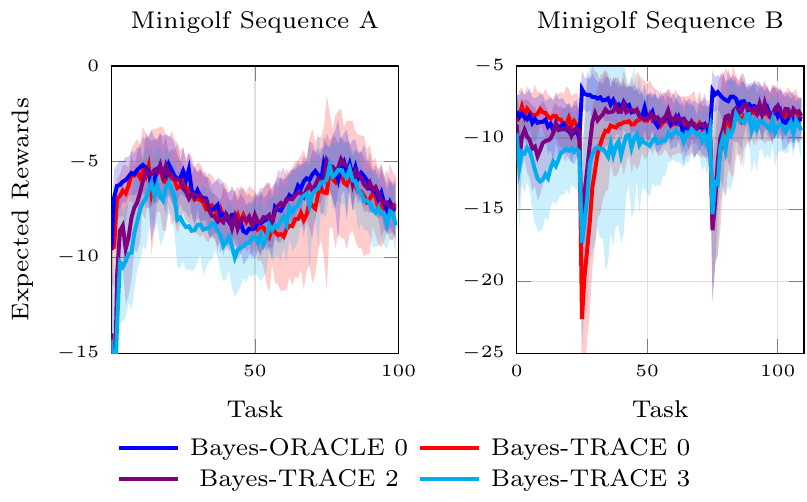}
\caption{Minigolf robustness experiment. We report Bayes-\algo and its oracle performances when different numbers of additional signals
are present. More specifically, the last number in the legend is the number of additional latent variables. Bayes-Oracle performances are reported when no additional latent variable is used.}
\label{fig:robust}
\end{figure*}

In this domain, tasks are specified by their goal position in 2D space (two latent variables). At meta-training, the latent variables are chosen in $[-3, 3]^{2}$. For both coordinates, our algorithm samples prior with means from $[-1, 1]$ and variances from $[0.1, 0.4]$. Once tasks are sampled, they are rescaled to the square $[-3, 3]^{2}$. 

The test sequences defined in Figure~\ref{fig:all} are: $x_{A}(t) = 3\sin({\frac{2\pi}{15}\sqrt{16t+5}})$, 
$y_{A}(t) = 3\cos({\frac{2\pi}{15}\sqrt{16t+5}})$; $x_{B}(t)= 3\sin{\frac{\pi}{4}}$ if $t \le 20 $ else $3\sin{\frac{5\pi}{4}}$ and $y_{B}(t)=3\cos{\frac{\pi}{4}}$ if $t \le 20 $ else $3\cos{\frac{5\pi}{4}}$. As initial prior, we use $\mathcal{N}([3\sin(0), 3\cos(0)], \mathrm{diag}(0.01, 0.01))$ for sequence A and $\mathcal{N}([x_{B}(0), y_{B}(0)], \mathrm{diag}(0.01, 0.01))$ for sequence B. Figure~\ref{fig:anty} shows tracking of the goal's Y-coordinate in these sequences, which was not reported in the main paper. Tasks are sampled for $x(t)$ and $y(t)$ rescaled to $[-1, 1]$ and variances $\mathrm{diag}(0.01, 0.01)$.

\subsection{Hyperparameters}

We use the PyTorch framework for our experiments. The main hyper-parameters can be found in Table \ref{tab:hyperparam}, while greater details can be found in our reference implementation (see code at \url{https://github.com/riccardopoiani/trio-non-stationary-meta-rl}).

\begin{table}
\centering
\begin{tabular}{llll}
\hline
Parameters  & Minigolf & HalfCheetahVel & AntGoal \\
\hline
Batch size     & 1280  & 6400 & 3200 \\
Epochs       & 4  & 2 & 2        \\
Minibatches  & 8  & 4 & 1 (B-\algo) 2 (TS-\algo)      \\
Clip paramer   & 0.1  & 0.1 & 0.1    \\
Max gradient norm & 0.5  & 0.5 & 0.5    \\
Entropy coefficient & 0  & 0.01 & 0.01    \\
Policy LR & 0.00005  & 0.0007 & 0.0005 \\
Inference LR & 0.2s  & 0.001 & 0.001    \\
Prior weight $\lambda$ & 1  & 0.1 & 0.1    \\
Policy number of layers & 2  & 2 & 2   \\
Policy unit per layers & 16  & 128 & 128 \\
Policy activation function & Tanh  & Tanh & Tanh \\

\hline
\end{tabular}
\caption{Main hyper-parameters used for training \algo's policy and inference networks.}
\label{tab:hyperparam}
\end{table}

\subsection{Additional Results}

\subsubsection{Robustness to latent variables}

In this experiment, we test the robustness of Bayes-\algo to useless latent variables, i.e., those that are not required to behave optimally. To do so, we add some useless state variables $x^{i}$ in the Minigolf domain that represent the distance between the current position $x$ of the ball and an unknown latent variable $\alpha^{i}$. More specifically, at each step, we have that $x^{i}_{t} = \lvert x - \alpha^{i} \rvert + \epsilon$ where $\epsilon \sim \mathcal{N}(0, 1)$. Then, we run Bayes-\algo on the same Minigolf domain with up to three useless states/latent-variables. Figure~\ref{fig:robust} reports the results on the same sequences used in the main paper where the additional latent variables $\alpha^{i}$ have, in each episode, a uniform random value in the range $[0, 2000\mathrm{cm}]$. We note that, in both cases, Bayes-\algo does not seem severely affected by the presence of multiple useless latent variables. This is an important result since, despite the inference network is trained to predict the distribution of \emph{every} latent variable, \algo's policy does not get distracted by these additional variables (i.e., it does not change its behavior to make the inference network better at predicting useless information). It is possible that even better performance could be achieved if we used a separate inference network for each latent variable.

\subsubsection{Additional Test Sequences}

\begin{figure*}[t]
\centering
\includegraphics[height=6.0cm]{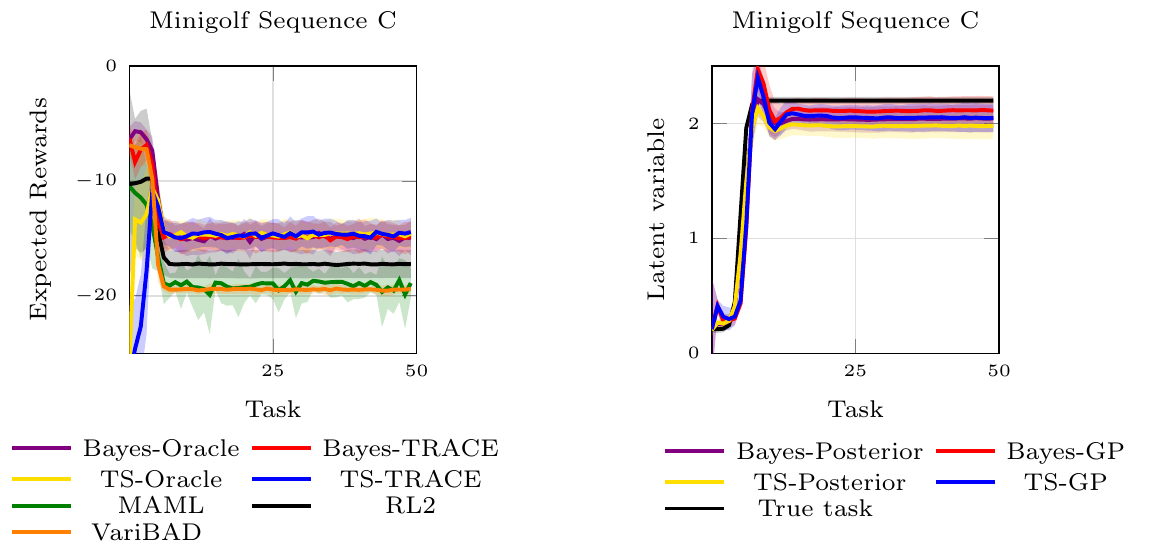} 
\caption{Meta-test performances on Minigolf Sequence C. Plots are 
averages and standard deviations of 20 policies, each of which is tested 50 times on the same episode; each task is composed of 4 episodes.}
\label{fig:additionalgolf}
\end{figure*}

\begin{figure*}[t]
\centering
\includegraphics[height=6.0cm]{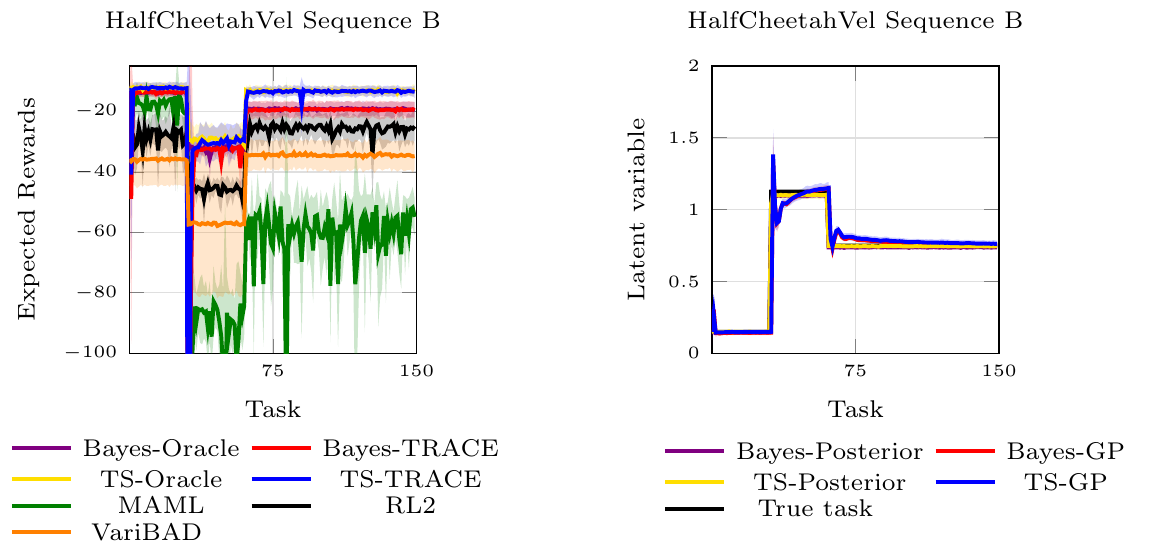} 
\caption{Meta-test performances on HalfCheetahVel Sequence B. Plots are 
averages and standard deviations of 5 policies, each of which is tested 50 times on the same episode; each task is composed of 1 episodes.}
\label{fig:additionalcheetah}
\end{figure*}

We report two additional test sequences for the experiments of the main paper in Figure~\ref{fig:additionalgolf} and ~\ref{fig:additionalcheetah}.

Minigolf sequence C is defined as $\omega_{C}(t)=0.995\tanh{(t-5)} + 1.204$. As initial prior we consider $\mathcal{N}(\rho_{C}(0), 0.2)$. We should note that, in the last steps of the sequence, the friction values are outside the training interval $[0.01, 2]$, and our method still outperforms the considered baselines. This verifies that \algo also generalizes out of the training distribution. Interestingly, RL2 at higher abrasion levels performs better than VariBAD, contrary to what happens when the friction coefficient is small.

HalfCheetahVel sequence B is defined as $v_{B}(t) = 0.15$ if $t \le 30$ else $1.125$ if $t \le 60$, $0.75$ otherwise. We use $\mathcal{N}(v_{B}(0), 0.00001)$ as initial prior for \algo. 
Interesting, we note that \algo suffers performance drops only when the first abprut change happens. At $t=60$ indeed, it is true that \algo believes that it needs to run at $1.125$, however, while increasing its velocity to reach the believed target speed, it is able to infer the correct latent space, thus avoiding further negative rewards. 
We also note that for speeds close to $0$ MAML has learned a good adaptation policy, comparable to the one of our oracles; however, when higher speeds are required, the policy suffers from poor local maxima.

\subsubsection{Regret plots}

\begin{figure*}[t]
\centering
\includegraphics[height=18.0cm]{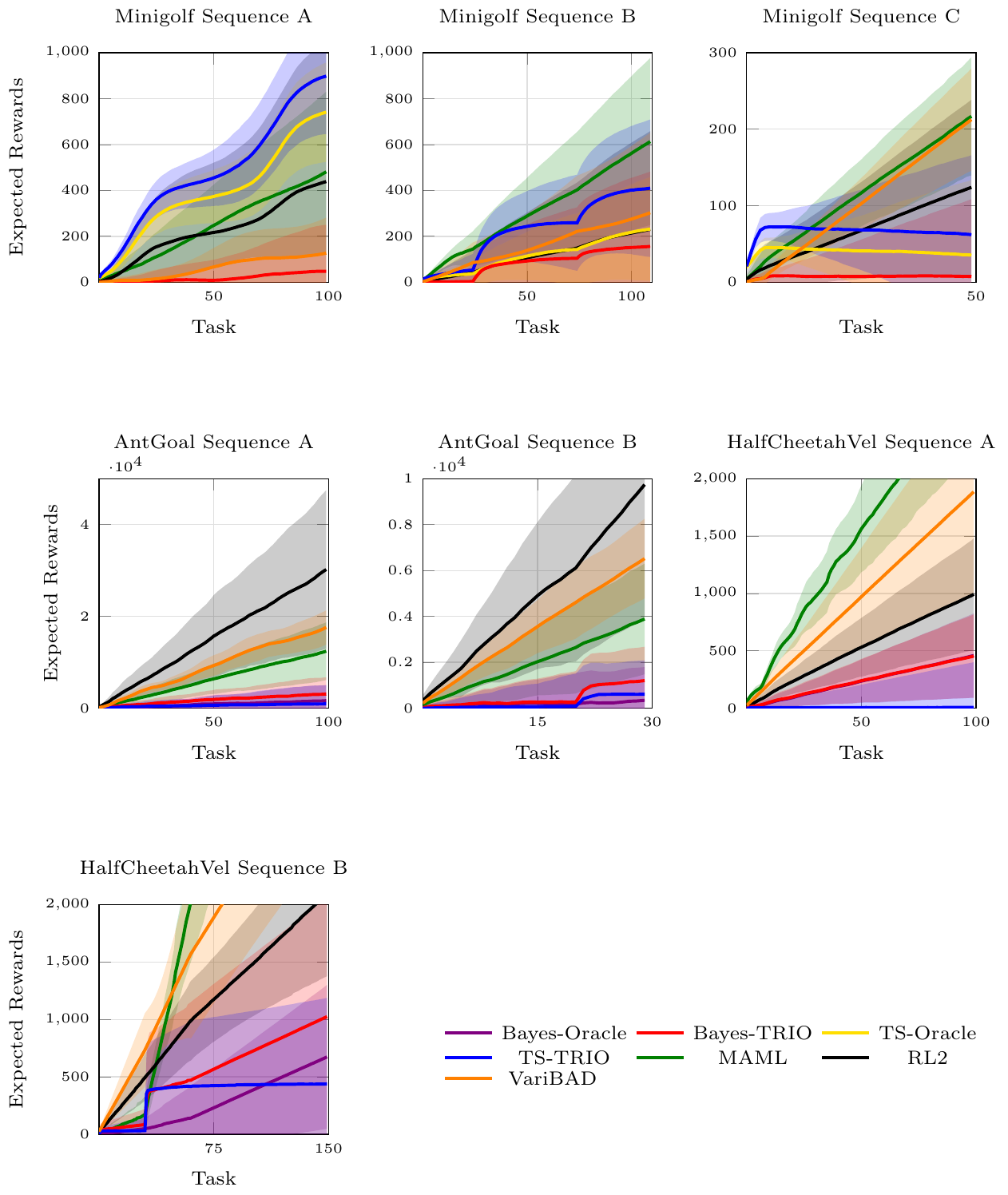} 
\caption{Meta-test regret on the presented sequences.}
\label{fig:regret}
\end{figure*}

To favor the comparison between the baselines considered in our experiments, here we report the plots of their regret at test time w.r.t. an oracle that always plays a near-optimal policy.
Formally, given an algorithm $A$ that receives return $r_t$ at episode $t$ and a clairvoyant algorithm $A^*$, receiving return $r^{*}_{t}$ at the same episode, the regret of $A$ over a sequence of tasks of length $T$ is defined as $R_{T}(A) := \sum_{t=1}^{T} (r_{t}^{*} - r_{t})$.
Figure ~\ref{fig:regret} reports the regret of the algorithms w.r.t. the best Oracle for the given problem. More specifically, we use Bayes-Oracle for Minigolf, and TS-Oracle for the MuJoCo experiments. 
 
\end{document}